%% file: cogsci_full_paper_template.tex
\documentclass[10pt,letterpaper]{article}

\usepackage{cogsci}

\cogscifinalcopy 

\usepackage[
  style=apa,
  natbib=true,
  annotation=false,
]{biblatex}
\usepackage{graphicx}
\usepackage{float} 
\usepackage{booktabs}
\usepackage{multirow}
\usepackage{array}
\usepackage{times}
\usepackage{helvet}
\usepackage{courier}
\usepackage[table]{xcolor}
\usepackage{subcaption}
\usepackage{algorithm}
\usepackage{algpseudocode}
\usepackage[strings]{underscore} 
\usepackage{enumitem} 
\usepackage{amsmath} 

\title{The Ratchet Effect in Silico: How Interaction Drives Cumulative Intelligence in Large Language Models}

\author[1]{\mbox{Ren Zhuang (venomrko97@gmail.com, 2022112011050@stu.hznu.edu.cn)}}
\affil[1]{School of Information Science and Technology, Hangzhou Normal University}

\begin{document}

\maketitle
\input{0_abstract.tex}
\input{1_intro.tex}
\input{2_related.tex}
\input{3_agora.tex}
\input{4_experiment.tex}
\input{5_discussion.tex}
\input{6_conclusion.tex}



\section{Acknowledgments}
I am grateful to Professor Ben Wang and Professor Shuifa Sun from Hangzhou Normal University for their guidance, encouragement, and continued support throughout the course of this work.

\printbibliography

\end{document}

%% file: 0_abstract.tex
\begin{abstract}
Human intelligence scales through cumulative cultural evolution (CCE), a ratchet process in which innovations are retained against entropic drift. Large language model training, by contrast, still depends primarily on static corpora and parameter growth, leaving little room for endogenous accumulation through interaction. We present POLIS (Population Orchestrated Learning and Inference Society), a framework in which heterogeneous agents generate solutions, verify one another's outputs, retain validated artifacts in shared cultural memory, and internalize them through parameter updates. On mathematical reasoning benchmarks, populations of 1--4B-parameter models achieved average gains of 8.8--18.9 points over base models and narrowed the gap to 70B+ monoliths. Mechanistic ablations identify peer verification as the main ratchet operator and show that internalization sustains accumulation across rounds, providing computational evidence that epistemic vigilance organizes durable knowledge growth. These results position structured social interaction as a scaling lever orthogonal to parameter count.

\textbf{Keywords} cumulative cultural evolution; ratchet effect; collective intelligence; epistemic vigilance; multi-agent learning
\end{abstract}

%% file: 1_intro.tex
\section{Introduction}

Human cognition scales through cumulative cultural evolution (CCE), and collective competence progressively exceeds the innovative capacity of any solitary individual \citep{mesoudi2018cumulative}. The ratchet effect that enables this depends on sociocognitive mechanisms ensuring innovations are selectively retained and transmitted with sufficient fidelity to prevent regression \citep{dean2012identification,tennie2009ratchet}. Unlike other primates exhibiting transient social learning \citep{whiten1999cultures,bandura1977social}, humans stabilize knowledge against entropic drift, transforming ephemeral interactions into durable cultural artifacts \citep{boyd2011cultural,henrich2015secret}. This scaling-by-organization is naturally framed as an emergent property of complex adaptive systems \citep{anderson1972more,lansing2003complex}.

Modern LLM training, by contrast, remains largely individualistic. Capability scales through parameters and static data \citep{kaplan2020scaling,hoffmann2022training}, with alignment imported from outside the agent's ecology \citep{ouyang2022training}. Despite emergent reasoning abilities \citep{wei2022emergent}, current models resemble isolated learners that exploit existing information but cannot sustain an autonomous ratchet effect, lacking the high-fidelity social transmission that CCE requires \citep{dean2012identification}.

\begin{figure}[t]
    \centering
    \includegraphics[width=.98\linewidth]{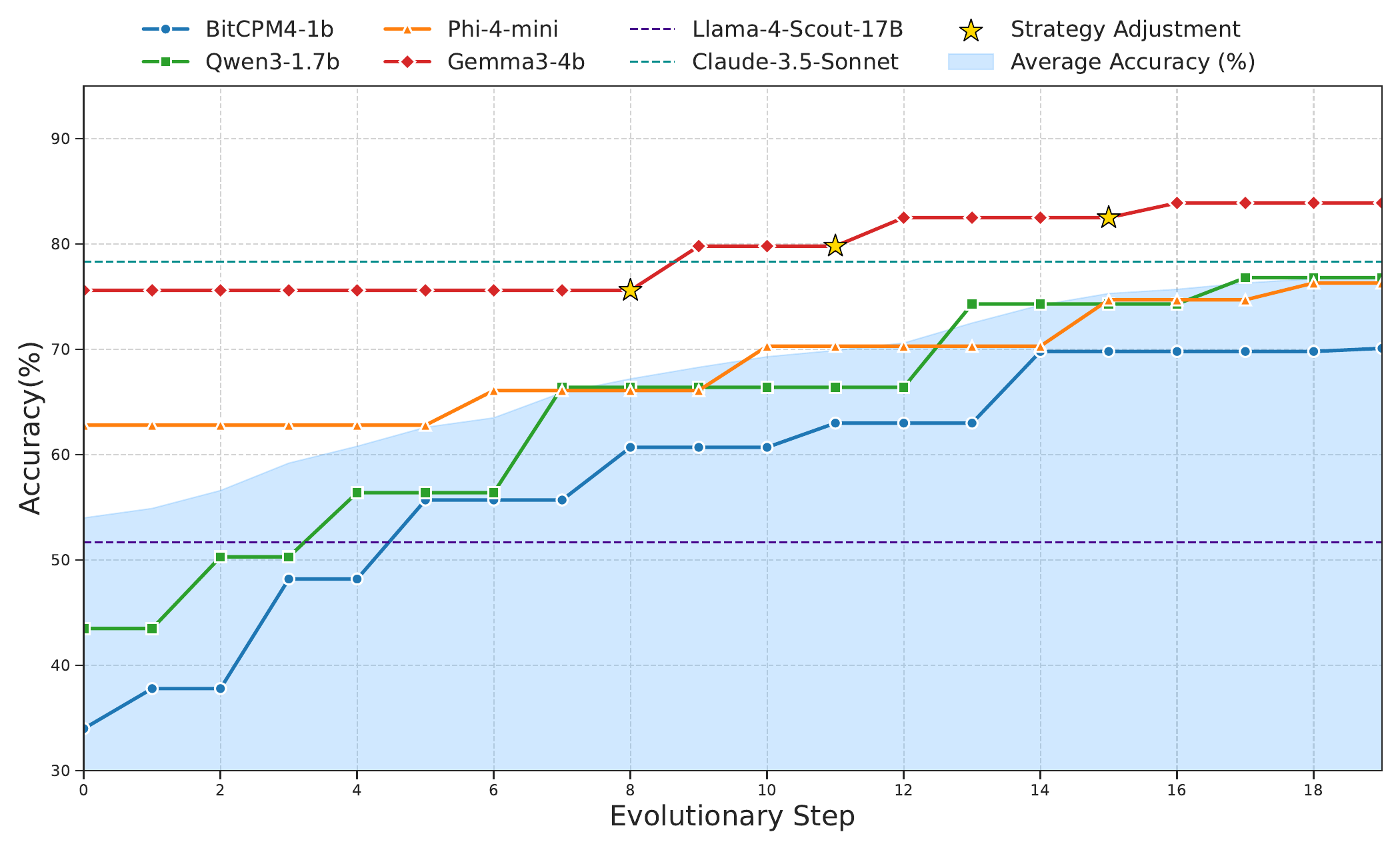}
    \caption{\textbf{Interaction supports cumulative gains.} POLIS improves over rounds, consistent with ratchet-like accumulation of verified strategies that reduces the gap to much larger monolithic models.}
    \label{fig:perf_evolution}
\end{figure}

We introduce \textbf{POLIS} (Population Orchestrated Learning and Inference Society), a computational framework for artificial sociogenesis that maps CCE mechanisms to language model training \citep{tomasello1999cultural}. POLIS organizes a heterogeneous micro-society in which agents generate solutions independently, verify one another's outputs, retain validated artifacts in shared cultural memory, and internalize them through parameter updates. This perspective aligns with calls to build machines that learn and think with people rather than scaling isolated learners alone \citep{collins2024building}, and connects to traditions on collective intelligence \citep{leimeister2010collective}. Our results show that structured social interaction can function as a scaling lever orthogonal to parameter count, as illustrated in Figure~\ref{fig:perf_evolution}, where small interacting models close the performance gap to much larger monoliths over rounds of CCE.

\paragraph{Our key contributions are}
\begin{itemize}
\item We formalize the ratchet cycle as an explicit loop over variation, social selection, and retention, implemented with modular operators for generation, verification, memory, and internalization.
\item We introduce a high-precision social learning architecture in which unanimous peer verification and preference-based quality selection admit transmissible artifacts into shared cultural memory.
\item We show across four reasoning benchmarks that this closed loop yields cumulative gains for heterogeneous populations, with verification and internalization governing sustained accumulation.
\end{itemize}

%% file: 2_related.tex
\section{Related Work}

\subsection{Cumulative Cultural Evolution and the Ratchet}
CCE explains open-ended cognitive scaling as a variation, selection, and inheritance cycle \citep{mesoudi2018cumulative,tomasello1999cultural} in which socially transmitted innovations are retained against loss \citep{dean2012identification}. The cognitive-gadgets hypothesis frames many reasoning procedures as culturally acquired software rather than fixed priors \citep{heyes2019precis}. Computational models of CCE often track transmission among agents or populations, whereas POLIS closes the loop between external artifacts and internal parameter updates. It implements the ratchet as an explicit training cycle in which candidate solutions are externalized, selectively retained through verification, and internalized into agent parameters so that gains persist across episodes.

\subsection{Epistemic Vigilance and Social Verification}
Interactionist accounts emphasize that evaluation is often more reliable than production \citep{mercier2011humans}, motivating epistemic vigilance mechanisms that filter misinformation \citep{csibra2009natural}. Collective decision-making improves accuracy when aggregation preserves independence rather than amplifying correlated noise \citep{bahrami2010optimally,woolley2010evidence}. Existing multi-agent LLM systems often aggregate outputs through majority voting or mixture-of-experts \citep{ebrahimi2025survey}. POLIS uses unanimous peer verification to decide what enters cultural memory, turning aggregation into a social selection process that prioritizes independent confirmation and suppresses correlated error.

\subsection{Multi-Agent Learning and LLM Self-Improvement}
LLM self-improvement methods such as STaR \citep{zelikman2022star} and related approaches \citep{li2023quantity,xu2024magpie,gao2025strategic,li2023mugglemath,tong2024dart} improve a single agent by generating and learning from its own solutions, so the selection signal remains local to that agent. Multi-agent finetuning introduces interaction for cross-validation \citep{chen2024multiagent}, and its effectiveness depends on preserving independence; homogeneous populations and low-precision aggregation can amplify correlated errors \citep{hong2004groups,lorenz2011social}. POLIS combines heterogeneous generation, peer verification, shared cultural memory, and parameter internalization in a single cross-round learning loop.

%% file: 3_agora.tex
\section{Modeling Artificial Sociogenesis}
\label{sec:method}

To realize a cumulative-cultural ratchet in artificial populations, a system must repeatedly generate novelty, select against error, and retain validated innovations with sufficient fidelity to prevent drift \citep{mesoudi2018cumulative,dean2012identification}. We formulate POLIS as a computational model of artificial sociogenesis that instantiates this cycle through adaptive scaffolding, decentralized verification, and parametric internalization.

\paragraph{Notation.}
We denote the heterogeneous population as $\mathcal{P} = \{ \pi_{\theta_i} \}_{i=1}^N$, parameterized by weights $\theta_i$.
The variation module comprises a stochastic generator $\mathcal{G}$, an adaptive scaffolding function $\mathcal{S}$, and an external anchor $\Omega$. Selection is mediated by an epistemic filter $\mathcal{F}$ and a quality preference $\mathcal{Q}$. Retention relies on a cultural memory $\mathcal{M}$ and an internalization process updating $\theta$.
Where needed, we denote the internalization step size by $\lambda_\theta$ to avoid ambiguity with LoRA hyperparameters.

\subsection{Variation Dynamics}
Evolution requires continuous variation \citep{boyd1988culture,hong2004groups}. We model it through three coupled mechanisms for generation, adaptation, and constraint.

\paragraph{Stochastic Generator ($\mathcal{G}$).}
To prevent mode collapse, problems are synthesized via a stochastic generator $\mathcal{G}$. Let $z \sim \mathcal{N}(0, I)$ be a latent noise vector. The problem set $Q_t$ is sampled conditioned on the current difficulty state $d_t$ as follows.
\begin{equation}
    q \sim \mathcal{G}(z \mid d_t) \quad \forall q \in Q_t.
\end{equation}

\paragraph{Adaptive Scaffolding ($\mathcal{S}$).}
To maintain the population within the \textit{zone of proximal development} (ZPD) \citep{van2010scaffolding}, the difficulty $d_t$ adapts homeostatically, closely related to curriculum learning in machine learning \citep{bengio2009curriculum}. Let $\bar{r}_t$ be the population average pass rate. Difficulty evolves as follows.
\begin{equation}
    d'_{t+1} \leftarrow d_t + \eta \cdot (\bar{r}_t - r_{\text{target}}),
\end{equation}
where $\eta$ is the adaptation rate.

\paragraph{External Anchor ($\Omega$).}
To prevent drift from external norms \citep{lorenz2011social}, we gate difficulty updates on a held-out generalization gap $\Delta_{\text{gen}}$ as follows.
\begin{equation}
    d_{t+1} = \begin{cases}
      d_t - \delta & \text{if } \Delta_{\text{gen}} \ge \Omega \\
      d_t + \eta(\bar{r}_t - r_{\text{target}}) & \text{otherwise}
    \end{cases}
\end{equation}

\subsection{Selection Dynamics}
The ratchet effect requires high-fidelity transmission \citep{dean2012identification}. We implement this via a two-stage filter.

\paragraph{Epistemic Filter ($\mathcal{F}$).}
First, we apply a decentralized truth-seeking mechanism. For a candidate solution $a$, a subset of peers $K$ acts as verifiers. Drawing on the asymmetry of argumentation \citep{mercier2011humans}, we require strict consensus, defined as
\begin{equation}
    \mathcal{F}(a) = \mathbb{I} \left[ \sum_{k \in K} \mathbb{I}(\text{Verify}_k(q, a)) = |K| \right].
\end{equation}
Solutions failing this check ($\mathcal{F}(a)=0$) are discarded. We set $K$ to all peers except the author, which imposes a conservative criterion and increases precision for suppressing hallucinations.

\paragraph{Quality Preference ($\mathcal{Q}$).}
Among verified solutions, evolution favors those that are communicable. Agents assign a quality score $\mathcal{Q}(a)$ (e.g., clarity), operationalized through peer preference judgments aggregated into a stable ranking signal such as TrueSkill \citep{herbrich2006trueskill}. The selected artifact $a^*$ is optimized for transmission as follows.
\begin{equation}
    a^* = \arg\max_{a \,\, \text{s.t.}\,\, \mathcal{F}(a)=1} \mathcal{Q}(a \mid q).
\end{equation}

\subsection{Retention Dynamics}
Finally, selected traits must be preserved to enable accumulation.

\paragraph{Cultural Memory ($\mathcal{M}$).}
We maintain an externalized buffer $\mathcal{M}_t$ representing the elite history of the society. The memory is updated via a priority queue mechanism as follows.
\begin{equation}
    \mathcal{M}_{t+1} \leftarrow \text{TopK}\left( \mathcal{M}_t \cup \{(q, a^*)\}, \mathcal{Q} \right).
\end{equation}
This buffer functions as an externalized scaffold \citep{kirsh1995intelligent}.

\paragraph{Internalization ($\theta$).}
The cycle closes when public artifacts are converted into private competencies. Consistent with the theory of cognitive gadgets \citep{heyes2019cognition}, agents update parameters $\theta$ via LoRA as follows.
\begin{equation}
    \theta_{t+1} \leftarrow \theta_t - \lambda_\theta \nabla_\theta \mathbb{E}_{(q,a) \sim \mathcal{M}_{t+1}} [\log \pi_\theta(a \mid q)].
\end{equation}

\subsection{Artificial Sociogenesis in POLIS}

Each interaction round follows a three-phase loop.
\begin{enumerate}
\item \textbf{Variation.} Generate challenges $q$ near the population's current competence in a ZPD-style regime \citep{vygotsky1978mind} using a stochastic generator $\mathcal{G}$. Adapt difficulty with scaffolding $\mathcal{S}$ and gate updates with an external regulator $\Omega$.
\item \textbf{Selection.} Agents propose solutions independently. Peers verify candidates via the epistemic filter $\mathcal{F}$ \citep{mercier2011humans}, and verified artifacts are ranked by quality preference $\mathcal{Q}$.
\item \textbf{Retention.} Selected artifacts enter cultural memory $\mathcal{M}$ and are internalized into each agent's parameters $\theta$ via lightweight updates (LoRA) \citep{hu2022lora}. This converts public solutions into private competence.
\end{enumerate}

%% file: 4_experiment.tex
\section{Experiments}
\label{sec:experiments}

We evaluated whether POLIS yields cumulative, ratchet-like improvements in mathematical reasoning and whether these gains depend on the specific sociocognitive operators defined in our model. Beyond final benchmark accuracy, we analyzed learning dynamics over interaction rounds and performed mechanistic ablations to identify which components are necessary to prevent drift.

\begin{table*}[t]
\centering
\small
\caption{\textbf{Social organization is a scaling axis.} POLIS consistently improves over both base models and isolated self-improvement baselines across benchmarks, supporting interaction structure as a complement to parameter scaling.}
\label{tab:model_performance}
\begin{tabular}{l|ccccc|ccccc}
\toprule
\textbf{Method} & \textbf{MATH500} & \textbf{GSM8K} & \textbf{GPQA} & \textbf{AIME24} & \textbf{Avg.} & \textbf{MATH500} & \textbf{GSM8K} & \textbf{GPQA} & \textbf{AIME24} & \textbf{Avg.} \\
\midrule
\textsc{Model} & \multicolumn{5}{c|}{\textsc{BitCPM4-1b}} & \multicolumn{5}{c}{\textsc{Qwen3-1.7b}} \\
\midrule
Baseline & 34.0 & 60.8 & 28.5 & 3.0 & 31.6 & 43.5 & 68.5 & 32.0 & 3.8 & 37.0 \\
STaR & 40.2 & 70.3 & 29.1 & 3.1 & 35.7 & 51.3 & 76.8 & 32.8 & 4.0 & 41.2 \\
SGDS & 54.5 & 78.1 & \underline{37.8} & 3.8 & 43.6 & 64.0 & 84.2 & 41.5 & 5.0 & 48.7 \\
MAGPIE & 39.8 & 72.4 & \textbf{40.1} & 3.0 & 38.8 & 49.5 & 78.0 & \underline{43.2} & 3.9 & 43.7 \\
GRA & 51.0 & 76.5 & 32.5 & 4.2 & 41.1 & 61.8 & 82.5 & 36.5 & 5.5 & 46.6 \\
MuggleMath & 46.5 & 79.5 & 30.2 & 3.4 & 39.9 & 58.5 & 85.1 & 34.5 & 4.4 & 45.6 \\
DART & \underline{57.1} & \underline{81.2} & 34.5 & \textbf{5.0} & \underline{44.5} & 66.2 & \underline{87.4} & 38.0 & \underline{6.1} & \underline{49.4} \\
\rowcolor[HTML]{EFEFEF} POLIS & \textbf{70.1} & \textbf{85.5} & 33.7 & \underline{4.9} & \textbf{47.4} & \textbf{76.8} & \textbf{90.3} & \textbf{49.1} & \textbf{7.5} & \textbf{55.9} \\
\midrule
\textsc{Model} & \multicolumn{5}{c|}{\textsc{Phi-4-mini}} & \multicolumn{5}{c}{\textsc{Gemma3-4b}} \\
\midrule
Baseline & 62.8 & 75.0 & 38.5 & 5.5 & 45.5 & 75.6 & 82.5 & 42.0 & 7.0 & 51.8 \\
STaR & 66.5 & 80.5 & 39.0 & 5.6 & 47.9 & 77.2 & 85.0 & 42.4 & 7.1 & 52.9 \\
SGDS & 70.8 & 85.5 & 44.0 & 6.2 & 51.6 & 80.1 & 87.0 & 46.5 & 7.9 & 55.4 \\
MAGPIE & 65.5 & 82.1 & \underline{45.5} & 5.5 & \underline{49.7} & 76.5 & 84.5 & \underline{48.1} & 7.0 & 54.0 \\
GRA & 70.1 & 84.0 & 41.0 & 6.5 & 50.4 & 79.5 & 86.8 & 44.0 & 8.0 & 54.6 \\
MuggleMath & 68.0 & 84.8 & 40.1 & 5.8 & 49.7 & 77.8 & 86.5 & 42.8 & 7.2 & 53.6 \\
DART & \underline{71.5} & \underline{86.5} & 42.5 & \underline{7.0} & \underline{51.9} & \underline{80.5} & \underline{88.2} & 45.0 & \underline{8.5} & \underline{55.6} \\
\rowcolor[HTML]{EFEFEF} POLIS & \textbf{76.3} & \textbf{89.2} & \textbf{47.4} & \textbf{8.1} & \textbf{55.3} & \textbf{83.9} & \textbf{91.8} & \textbf{56.5} & \textbf{10.2} & \textbf{60.6} \\
\bottomrule
\end{tabular}
\end{table*}

\begin{figure*}[t]
\centering
\begin{subfigure}[b]{0.345\textwidth}
    \centering
    \caption{Gains with rising latency}
    \label{fig:perf_time}
    \includegraphics[width=\textwidth]{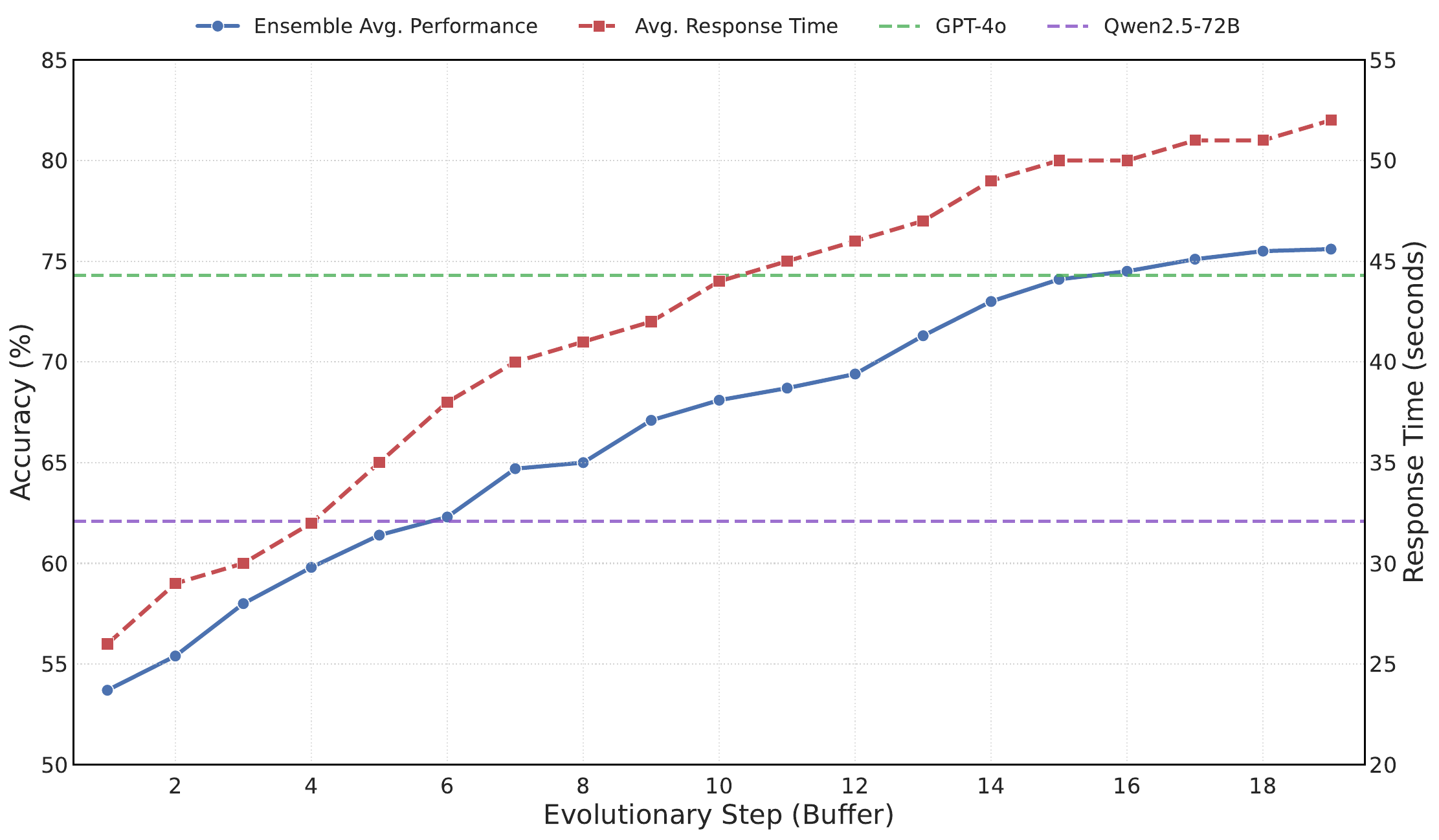}
\end{subfigure}
\hfill
\begin{subfigure}[b]{0.31\textwidth}
    \centering
    \caption{Diversity accelerates accumulation}
    \label{fig:group_evolution}
    \includegraphics[width=\textwidth]{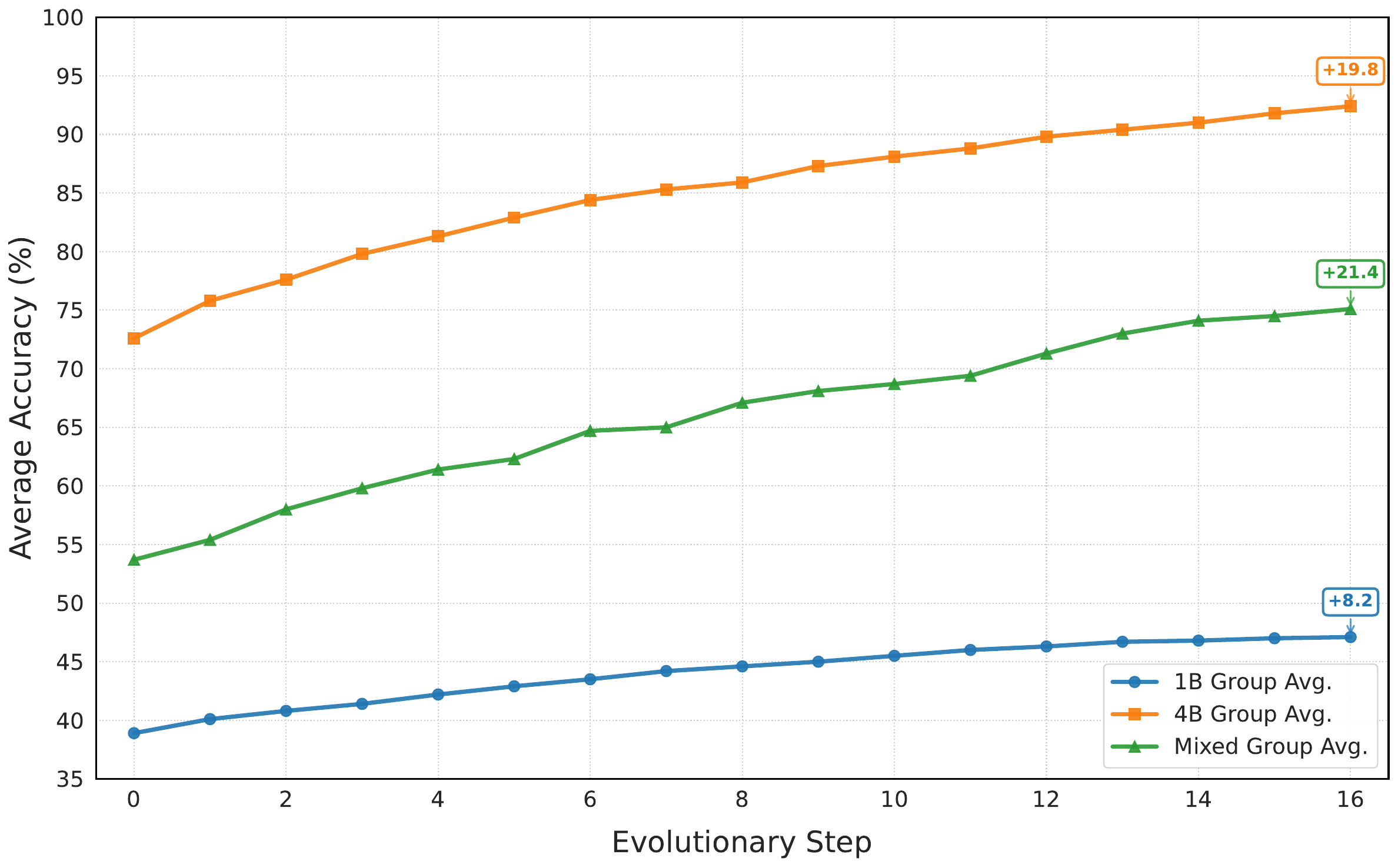}
\end{subfigure}
\hfill
\begin{subfigure}[b]{0.335\textwidth}
    \centering
    \caption{Scaffolding raises artifact yield}
    \label{fig:efficiency_dynamics}
    \includegraphics[width=\textwidth]{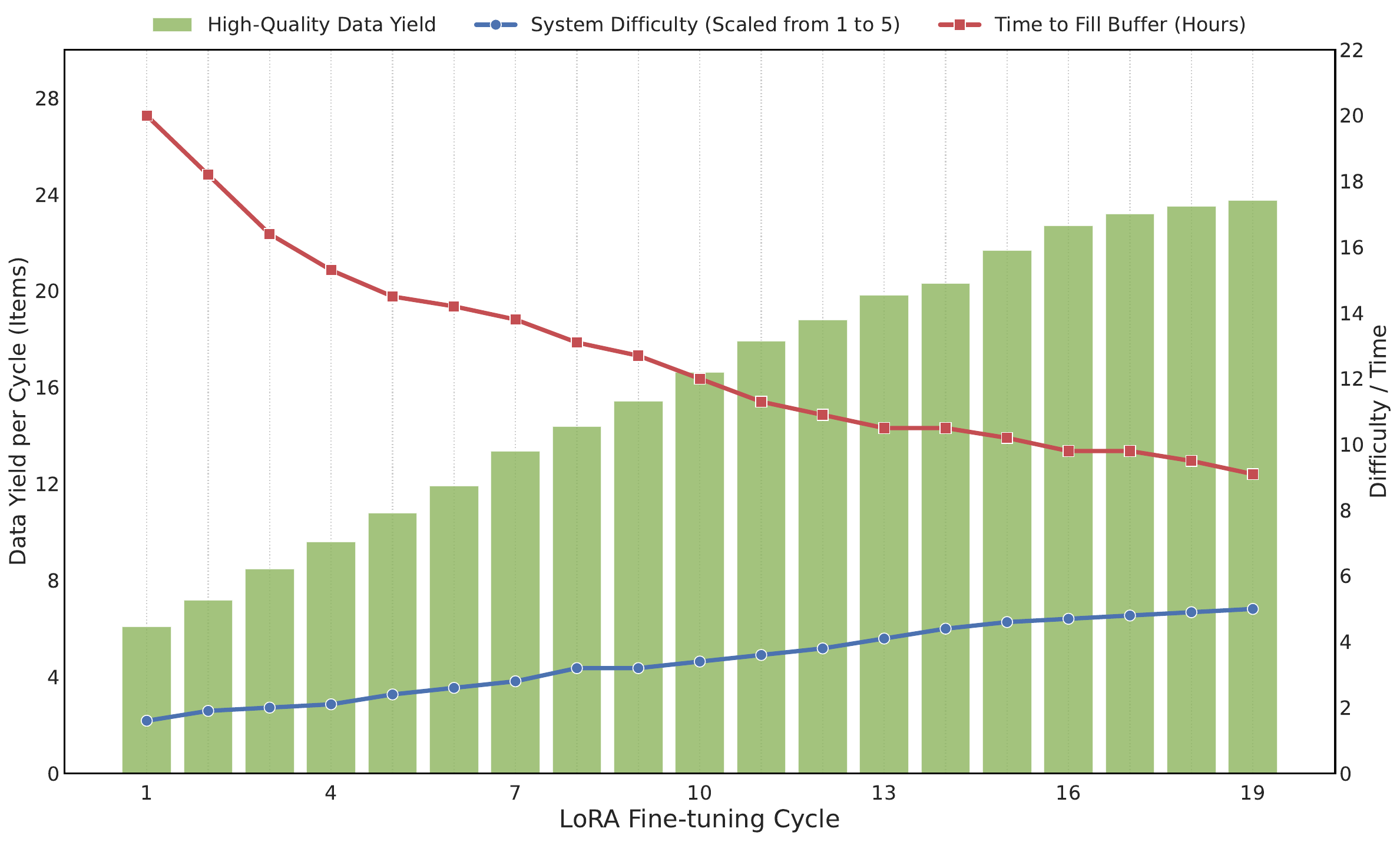}
\end{subfigure}
\caption{\textbf{Population dynamics and trade-offs.}
POLIS exhibits ratchet-like accumulation across rounds, and this accumulation is accelerated by diversity and supported by adaptive scaffolding that increases the yield of verified training artifacts while improving efficiency.}
\label{fig:combined_results}
\end{figure*}

\subsection{Experimental Setup}
\label{sec:experiments_setup}

\paragraph{Models.}
We form a heterogeneous population of four LLMs ranging from 1B to 4B parameters, including BitCPM4 \citep{team2025minicpm4}, Qwen3 \citep{yang2025qwen3}, Phi-4 \citep{abouelenin2025phi}, and Gemma3 \citep{team2025gemma}, Llama3.2 \citep{dubey2024llama}.

\paragraph{Benchmarks.}
We evaluate on MATH500 \citep{hendrycks2021measuring}, GSM8K \citep{cobbe2021training}, GPQA Diamond \citep{rein2024gpqa}, and AIME24 \citep{AIME} using Chain-of-Thought (CoT) prompts \citep{wei2022chain}. We focus on mathematical reasoning as its clear ground truth enables unsupervised decentralized verification, and its multi-step structure renders cumulative strategy acquisition directly observable. This setting makes social filtering and cross-round retention directly measurable within the learning loop.

\paragraph{Baselines.}
We compare POLIS against isolated self-improvement methods, including STaR \citep{zelikman2022star}, SGDS \citep{li2023quantity}, MAGPIE \citep{xu2024magpie}, GRA \citep{gao2025strategic}, MuggleMath \citep{li2023mugglemath}, DART \citep{tong2024dart}, along with monolithic systems such as GPT-4o \citep{hurst2024gpt}, Claude 3.5 Sonnet \citep{anthropic2024claude}, Qwen2.5-72B \citep{team2024qwen2}, and Llama-4-Scout-17B \citep{Llama4}.

\paragraph{Implementation.}
Internalization updates parameters via LoRA ($r{=}16$, $\alpha{=}32$, dropout $0.05$) \citep{hu2022lora} after 1024 verified artifacts accumulate. Adaptive scaffolding initializes at $d_0{=}1.0$ with update rate $0.05$; generation uses temperature $0.6$ and top-$p$ $0.95$.

\subsection{Main Results}

\paragraph{Social Organization as a Scaling Axis.}
Table \ref{tab:model_performance} presents the primary comparative results across four benchmarks. POLIS consistently improves each base model and outperforms strong isolated self-improvement baselines. The baseline set spans local self-training, strategic data generation, and larger monolithic models, positioning POLIS within a broad map of established improvement regimes. Averaged across tasks, POLIS yields gains of 8.8--18.9 points over the base models and 2.9--6.5 points over the strongest isolated baseline among those tested. Interaction structure thus emerges as a scaling axis alongside parameter count. MAGPIE slightly outperforms POLIS on GPQA with smaller models; we attribute this to its emphasis on diverse instruction generation, which aligns well with the benchmark's graduate-level question variety. Nevertheless, POLIS surpasses MAGPIE on GPQA with larger models and achieves higher average accuracy across all benchmarks.

\begin{figure*}[t]
\centering
\includegraphics[width=.98\textwidth]{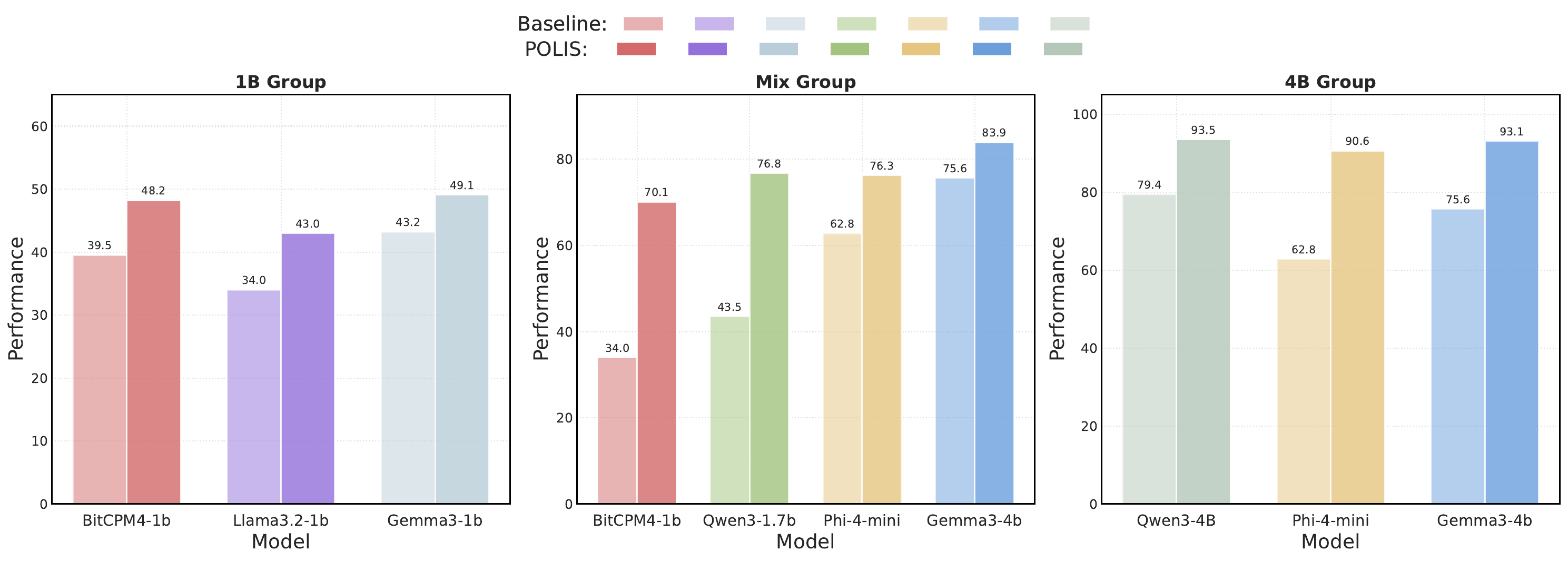}
\caption{\textbf{Diversity redistributes gains without sacrificing strong models.} POLIS improves every member of the population, with the largest gains accruing to weaker models in heterogeneous societies while stronger models do not regress.}
\label{fig:model_perf_comp}
\end{figure*}

\paragraph{Verifying the ratchet effect.}
Figure~\ref{fig:perf_time} shows that population performance accumulates steadily over rounds, a hallmark of CCE \citep{dean2012identification}. The curve exhibits a consistent upward trajectory rather than transient fluctuations, while average response time increases with difficulty as agents internalize longer strategies.

\paragraph{Cognitive Diversity and Collective Intelligence.}
We investigate whether gains stem from population structure rather than agent count alone. Figure \ref{fig:group_evolution} contrasts the Mixed Group against scale-homogeneous controls. The heterogeneous society improves faster and reaches a higher plateau, consistent with the diversity prediction theorem \citep{hong2004groups} and with heterogeneity reducing correlated error in homogeneous settings \citep{lorenz2011social}.
More broadly, these dynamics match classic accounts of collective intelligence as an organizational resource rather than a property of any single agent \citep{leimeister2010collective,hutchins1995cognition}.

\paragraph{Efficiency dynamics.}
Figure~\ref{fig:efficiency_dynamics} tracks system metabolism, defined as artifact yield relative to problem difficulty.
Even as difficulty rises, the yield of verified artifacts increases, indicating that the population internalizes strategies rather than merely memorizing answers.

\subsection{Analysis}

\paragraph{Ablation Study.}
Cumulative evolution depends on selective retention of variants against noise. Ablation studies of our model's sociocognitive mechanisms, as illustrated in Table~\ref{tab:ablation_study}, show that removing the Epistemic Filter causes the largest performance drop, consistent with argumentative theory \citep{mercier2011humans}. The filter governs which artifacts enter cultural memory and therefore which reasoning traces are internalized across rounds. Ablating the Quality Preference also impairs performance, revealing that the Epistemic Filter enforces correctness while the Quality Preference enhances communicability. The system is more impaired by a lack of neural plasticity than by a lack of historical records, supporting the view that external artifacts serve as temporary scaffolds whose durable effect arises through parameter updates \citep{heyes2019precis}.

\begin{table}[t]
\centering
\small
\caption{\textbf{Verification is the main ratchet operator.} Removing the Epistemic Filter yields the largest drop in average accuracy, while internalization and quality selection also contribute to sustained accumulation.}
\label{tab:ablation_study}
\begin{tabular}{lc}
\toprule
\textbf{Method Variant} & \textbf{Avg. Acc} \\
\midrule
POLIS & 54.8 \\
- Epistemic Filter $\mathcal{F}$      & 37.8 \\
- Quality Preference $\mathcal{Q}$   & 47.8 \\
- Internalization $\theta$            & 45.5 \\
- Cultural Memory $\mathcal{M}$       & 49.6 \\
- Adaptive Scaffolding $\mathcal{S}$ & 53.6 \\
- Stochastic Generator $\mathcal{G}$ & 52.8 \\
- External Anchor $\Omega$           & 53.3 \\
\bottomrule
\end{tabular}
\end{table}

\begin{figure*}[t]
\centering
\includegraphics[width=.98\textwidth]{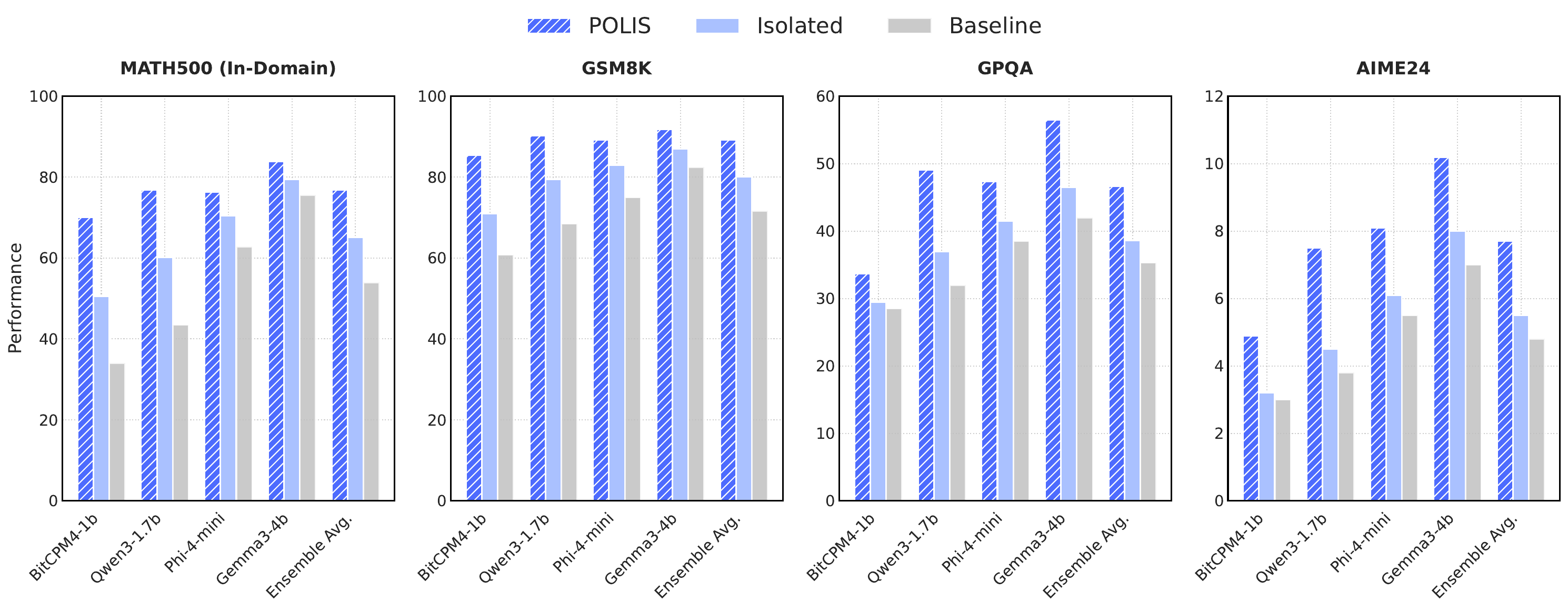}
\caption{\textbf{Interaction yields robust gains beyond solitary learning.} POLIS improves over both base models and an isolated-learning control, consistent with peer verification reducing errors that escape single-agent verification horizons.}
\label{fig:multi_dataset_perf_comp}
\end{figure*}

\paragraph{The Dividend of Cognitive Diversity.}

Figure \ref{fig:model_perf_comp} isolates the effect of population structure by comparing models trained in homogeneous groups versus the heterogeneous POLIS society. Diversity functions as a non-rivalrous resource: weaker agents benefit most, stronger agents retain performance, and the group preserves complementary strengths under shared verification. This pattern matches the diversity prediction theorem \citep{hong2004groups} and aligns with distributed cognition accounts of collective performance \citep{hutchins1995cognition,lorenz2011social}.

\paragraph{Comparing Social and Solitary Generalization.}

Figure \ref{fig:multi_dataset_perf_comp} contrasts the social trajectory against isolated learning across four distinct cognitive environments. Across all benchmarks, POLIS outperforms both the base models and the isolated-learning control. On AIME24, for example, BitCPM4-1b improves by 63\% relative to baseline, whereas isolated learning yields only a modest gain. The gap reflects a structural property of the learning loop: peer verification governs which artifacts are retained and internalized, improving generalization across regimes.

%% file: 5_discussion.tex
\section{Discussion}
\label{sec:discussion}

\subsection{Interaction as a Scaling Law}
POLIS shows that capability can scale through interaction structure as well as parameter count, consistent with distributed and situated accounts of cognition \citep{hollan2000distributed,hutchins1995cognition,brown1989situated}. In our framework, decentralized verification establishes an endogenous epistemic norm by granting candidate artifacts knowledge status only after independent peer checking, aligning the production-evaluation asymmetry in interactionist theories of reasoning \citep{mercier2011humans} with metacognitive monitoring \citep{flavell1979metacognition}. Ablation results reveal a functional hierarchy among operators. Removing the Epistemic Filter causes the greatest performance decline, followed by internalization and Quality Preference. Variation supplies candidates; social filtering determines which ones persist \citep{tennie2009ratchet}.

\subsection{The Cultural Ratchet}
The gains in POLIS depend on both selection and retention. Unanimous verification and quality-based preference preserve artifacts that are both correct and communicable, while internalization converts these public artifacts into private competence, aligning with the idea that many reasoning procedures function as culturally learned cognitive gadgets \citep{heyes2019precis} and with the broader cultural intelligence hypothesis \citep{herrmann2007humans,tomasello2005understanding}. This sequence of externalization and internalization reflects how social interaction scaffolds individual development \citep{vygotsky1978mind}. In POLIS, the cultural memory $\mathcal{M}$ serves as a zone of proximal development by providing exemplars just beyond each agent's current competence. The observation that weaker models benefit disproportionately in heterogeneous societies, as shown in Figure~\ref{fig:model_perf_comp}, supports this scaffolding interpretation.

\subsection{Implications for Human CCE}
POLIS offers a computational testbed for probing the foundations of cumulative cultural evolution and, more specifically, a micro-institution for cumulative intelligence. Our results isolate three ingredients for this process, namely mechanisms that generate diversity, high-precision social filters that reject errors, and pathways that internalize innovations into stable competence. The prominence of verification aligns with anthropological evidence that cumulative cultures depend strongly on population connectivity and error-correction norms \citep{derex2016partial,powell2009late}. Human epistemic vigilance remains richer, drawing on mental-state reasoning \citep{frith2006neural,heyes2014cultural}, whereas POLIS agents rely on outcome-based consensus. This contrast makes POLIS useful for cognitive science in a second sense. It separates innovation, social evaluation, memory, and internalization into distinct operators that can be manipulated experimentally. Population structure, verifier independence, and memory policy can therefore be studied as organizational variables rather than treated as background conditions. Extending this framework toward richer social inference is a natural next step for computational studies of CCE.

\subsection{Scalability and Trade-offs}
POLIS highlights critical design tensions in multi-agent societies. Verification improves information quality but raises interaction costs at scale, making verifier allocation and communication topology central variables. Internalization, conversely, risks converging agents toward local optima and eroding diversity. Effective societies therefore require sufficient structural commonality to stabilize transmission, paired with enough heterogeneity to preserve complementary strengths. These dynamics mirror organizational learning: institutions standardize evaluation for reliability while sustaining distributed perspectives for adaptability. POLIS similarly clarifies the cultural ratchet effect, where cumulative progress hinges on retaining error correction without depleting the generative diversity needed for novel variants. Within the framework, this balance is governed by verifier independence, memory selectivity, and update frequency.

By operationalizing these tensions, POLIS shifts the research focus from static accuracy benchmarks to investigate which organizational structures maximize long-term knowledge accumulation under fixed compute and communication budgets. Treating institutions rather than isolated agents as the primary unit of analysis bridges cognitive science and machine learning. Furthermore, the framework facilitates controlled ablation studies that systematically compare centralized and decentralized verification protocols, evaluate sparse and dense communication topologies, and examine varying memory horizons, enabling a rigorous exploration of social architecture under constrained resources.

%% file: 6_conclusion.tex
\section{Conclusion}
\label{sec:conclusion}

The dominant approach to scaling intelligence trains ever larger models on expanding datasets. POLIS shows how populations of bounded reasoners accumulate capability through social interaction. Experiments identify verification as the operator that governs what the society retains, while internalization turns shared artifacts into durable competence. This result aligns with cognitive science accounts in which epistemic vigilance stabilizes cumulative learning. Viewing intelligence as scalable through organization shifts the question from model size to connectivity. POLIS provides a testbed for this scaling axis and for institutional variation in cumulative intelligence.